\documentclass[10pt,twocolumn]{article} 
\usepackage{simpleConference}
\usepackage{times}
\usepackage{graphicx}
\usepackage{amssymb}
\usepackage{gensymb}
\usepackage{amsmath}
\usepackage{url}
\usepackage{hyperref} 
\usepackage{breakurl}
\def\UrlBreaks{\do\/\do-}
\hypersetup{
colorlinks,
linkcolor=blue,
citecolor=blue,
filecolor=blue,
urlcolor=blue}
\usepackage[labelfont={bf},]{caption}
\usepackage[none]{hyphenat}

\usepackage{multirow}

\usepackage{listings}
\usepackage{color}

\usepackage[noadjust]{cite}

\definecolor{mygreen}{rgb}{0,0.6,0}
\definecolor{mygray}{rgb}{0.5,0.5,0.5}
\definecolor{mymauve}{rgb}{0.58,0,0.82}

\newcommand{\longurl}[1]{%
    {\expandafter\def\expandafter\UrlBreaks\expandafter{\UrlBreaks\UrlOrds%
        \do\/\do\a\do\b\do\c\do\d\do\e\do\f%
        \do\g\do\h\do\i\do\j\do\k\do\l\do\m%
        \do\n\do\o\do\p\do\q\do\r\do\s\do\t%
        \do\u\do\v\do\w\do\x\do\y\do\z%
        \do\A\do\B\do\C\do\D\do\E\do\F\do\G%
        \do\H\do\I\do\J\do\K\do\L\do\M\do\N%
        \do\O\do\P\do\Q\do\R\do\S\do\T\do\U%
        \do\V\do\W\do\X\do\Y\do\Z}%
    \url{#1}}%
}

\usepackage{etoolbox}
\makeatletter
\patchcmd\@combinedblfloats{\box\@outputbox}{\unvbox\@outputbox}{}{\errmessage{\noexpand patch failed}}
\makeatother

\newcommand\extraspace{3pt}

\begin{document}

\title{Improving Electron Micrograph Signal-to-Noise with an Atrous Convolutional Encoder-Decoder}

\author{Jeffrey M. Ede \\
\\
j.m.ede@warwick.ac.uk  \\
}

\maketitle

\noindent \textbf{Abstract:} We present an atrous convolutional encoder- decoder trained to denoise 512$\times$512 crops from electron micrographs. It consists of a modified Xception backbone, atrous convoltional spatial pyramid pooling module and a multi-stage decoder. Our neural network was trained end-to-end to remove Poisson noise applied to low-dose ($\ll 300$ counts ppx) micrographs created from a new dataset of 17267 2048$\times$2048 high-dose ($>$ 2500 counts ppx) micrographs and then fine-tuned for ordinary doses (200-2500 counts ppx). Its performance is benchmarked against bilateral, non-local means, total variation, wavelet, Wiener and other restoration methods with their default parameters. Our network outperforms their best mean squared error and structural similarity index performances by 24.6\% and 9.6\% for low doses and by 43.7\% and 5.5\% for ordinary doses. In both cases, our network's mean squared error has the lowest variance. Source code and links to our new high-quality dataset and trained network have been made publicly available at \url{https://github.com/Jeffrey-Ede/Electron-Micrograph-Denoiser}.



\section{Introduction}
\noindent Every imaging mode in electron microscopy is limited by and has been shaped by noise\cite{oho1996practical}. Increasingly, ever more sophisticated and expensive hardware and software based methods are being developed to increase resolution, including aberration correctors\cite{pennycook2017impact, linck2016chromatic}, advanced cold field emission guns\cite{houdellier2015development, akashi2015aberration}, holography\cite{adaniya2018development, koch2014towards} and others\cite{feist2017ultrafast, migunov2015rapid, jiang2018electron}. However, these developments are all fundamentally limited by the signal-to-noise ratios in the micrographs they are being applied to. Low-dose applications such as single-particle cryogenic microscopy\cite{hattne2018analysis} and real-time tomography\cite{migunov2015rapid} are also complicated, made preventively difficult or limited by noise.

Moving towards higher resolution, a large number of general\cite{motwani2004survey} and electron microscopy-specific\cite{oho1996practical, zhang2010cryo} denoising algorithms have been developed. However, most of these algorithms rely on laboriously hand-crafted filters and are rarely; if ever, truly optimized for their target domains e.g. \cite{kushwaha2012noising}. Neural networks are universal approximators\cite{hornik1989multilayer} that overcome these difficulties\cite{lin2017does} through representation learning\cite{bengio2013representation}. As a result, networks are increasingly being applied to noise removal\cite{yang2018low, remez2017deep, mao2016image, zhang2017beyond} and other applications in electron microscopy\cite{xu2017deep, lee2017superhuman, ciresan2012deep, zhu2017deep} and other areas of science. 


The recent success of large neural networks in computer vision is attributed to the advent of graphical processing unit (GPU) acceleration\cite{schmidhuber2015deep, chetlur2014cudnn}. In particular, the GPU acceleration of large convolutional neural networks\cite{liu2017survey, mccann2017review} (CNNs) in distributed settings\cite{chen2016revisiting, abadi2016tensorflow}. Large networks that surpass human performance in image classification\cite{christiansen2018silico, he2015delving}, computer games\cite{mnih2013playing, lample2017playing, firoiu2017beating}, speech recognition\cite{xiong2016achieving, weng2014single}, relational reasoning\cite{santoro2017simple} and in many other applications\cite{lee2017superhuman, wang2018deep, yu2017sketch, han2018deep, weyand2016planet} have all been enabled by GPUs. 

At the time of writing, there are no large neural networks for electron micrograph denoising. Instead, most denoising networks act on many small overlapping crops e.g. \cite{mao2016image}. This makes them computationally inefficient and unable to utilize all the information available. Some large denoising networks have been trained as part of generative adversarial networks\cite{yang2018low_paper2} and try to generate micrographs resembling high-quality training data as closely as possible. This can avoid the blurring effect of most filters; however, this is achieved by generating features that might be in high-quality micrographs. This means that they are prone to producing artifacts; something that is often undesirable in scientific applications.

This paper presents a large CNN for electron micrograph denoising. Network architecture is shown in section~\ref{sec_architecture}. The collation of a new high-quality electron micrograph training dataset, training hyperparameters and learning protocols are described in section~\ref{sec_training}. Performance comparison with other methods', error characterization and example usage are in section~\ref{sec_performance}. Finally, architecture and hyperparameter tuning experiments are presented in section~\ref{additional_experiments}.

\begin{figure*}[tbp]
\centering
\includegraphics[width=0.95\textwidth]{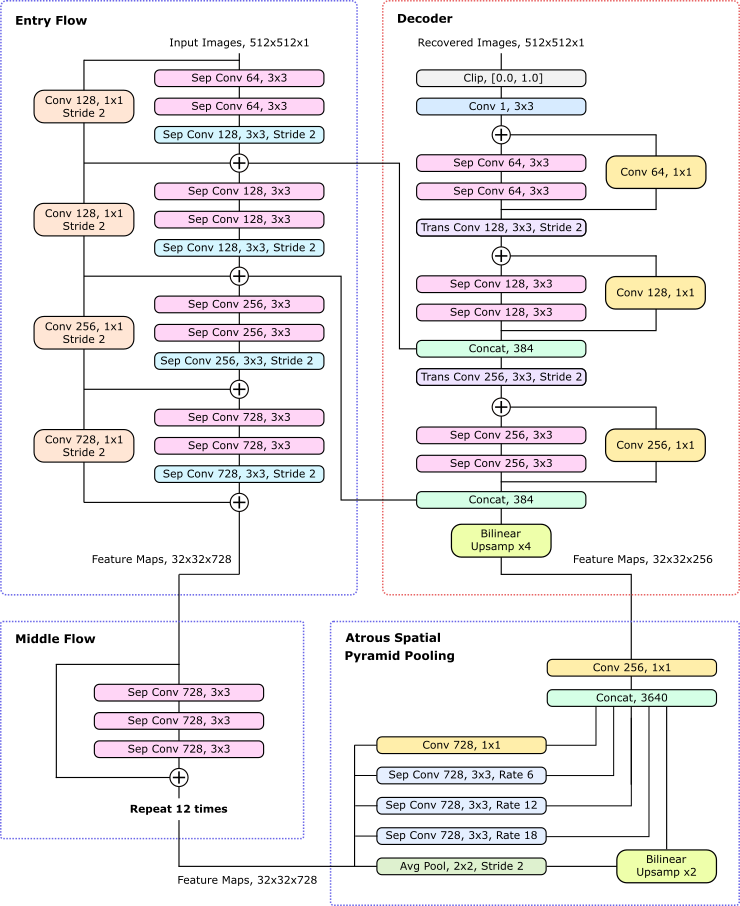}
\vspace{.03cm}
\caption{ Architecture of our deep convolutional encoder-decoder for electron micrograph denoising. The entry and middle flows develop high-level features that are sampled at multiple scales by the atrous spatial pyramid pooling module. This produces rich semantic information that is concatenated with low-level entry flow features and resolved into denoised micrographs by the decoder.} 
\label{noise-removal-nn}
\end{figure*}

\section{Architecture}\label{sec_architecture}
\noindent Our highest performing denoising network was trained for 512$\times$512 inputs and is shown in fig.~\ref{noise-removal-nn}. It consists of modified Xception\cite{chollet2016xception} entry and middle flows for feature extraction, an atrous spatial pyramid pooling (ASPP) module\cite{chen2017rethinking, chen2018encoder} that samples rich high-level semantics at multiple scales and a multi-stage decoder that combines low-level entry flow features with ASPP semantics to resolve them. The architecture is inspired by Google's DeepLab3\cite{chen2017rethinking}, DeepLab3+\cite{chen2018encoder} and other encoder-decoder architectures\cite{yang2018low, badrinarayanan2017segnet, mao2016image}. A guide for readers unfamiliar with convolution arithmetic for deep learning is \cite{dumoulin2016guide}.

Convolutions all use 3$\times$3 or 1$\times$1 kernels and are followed by batch normalization\cite{ioffe2015batch} before ReLU6\cite{krizhevsky2010convolutional} activation. Similar to DeepLab3+, original Xception max pooling layers have been replaced with strided depthwise separable convolutions, enabling the network to learn its own downsampling. Extra batch normalization is added between the depthwise and pointwise convolutions of every depthwise separable convolution like in MobileNet\cite{howard2017mobilenets}.

The following subsections correspond to the subsections of fig.~\ref{noise-removal-nn}. Training details follow in section~\ref{sec_training}.

\subsection{Entry Flow}

\begin{figure}[tbp]
\centering
\includegraphics[width=0.36\textwidth]{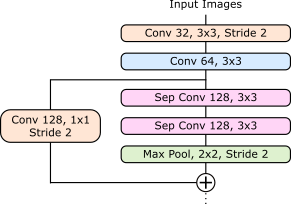}
\caption{ Start of an unmodified Xception entry flow\cite{chollet2016xception}. }
\label{xception_start}
\end{figure}

\noindent Other than the modification of max pooling layers to strided depthwise separable convolutions, most entry flow convolutions are similar to Xception's\cite{chollet2016xception}. Common convolutions have the same number of features and are arranged into Xception-style residual\cite{he2016deep} blocks to reduce semantic decimation during downsampling\cite{chen2017rethinking}.

The main change is the replacement of the first two Xception convolutions with an residual convolutional downsampling block. For reference, the start of the original Xception entry flow is shown in fig.~\ref{xception_start}. This change was made so that the lowest-level features concatenated in the decoder would be deeper and have a larger feature space. An alternative solution is to apply additional pre-concatenation 1$\times$1 convolutions to change the feature space size. This was the approach taken in DeepLab3+\cite{chen2018encoder} to prevent high-level ASPP semantics being overwhelmed by low-level features in the decoder.

\subsection{Middle Flow}

\noindent Skip-3 residual blocks are repeated 12 times to develop high-level semantics to flow into the ASPP module. This is more than the 8 skip-3 blocks used in Xception\cite{chollet2016xception} as the inflowing tensors are larger; 32$\times$32$\times$728 rather than 19$\times$19$\times$728. Nevertheless, this is fewer than the 16 skip-3 residual blocks used in DeepLab3+ for similar tensors\cite{chen2018encoder}. Middle flow tensors are much smaller than those in other parts of the network so this decision was a compromise between expressibility and training time.

\subsection{Atrous Spatial Pyramid Pooling}

\noindent This is the ASPP module Google developed for semantic image segmentation\cite{chen2017rethinking, chen2018encoder} without a pre-pooling 1$\times$1 convolution. It is being used rather than a fully connected layer; like the one in \cite{yang2018low}'s denoiser, to sample semantics at multiple scales as it requires fewer parameters and has almost identical performance. The atrous rates of 6, 12 and 18 are the same as those used in the original ASPP module\cite{chen2017rethinking}. Importantly, a bottleneck 1$\times$1 convolution is used to reduce the number of output features from 3640 to 256, forcing the network to work harder to develop high-level semantics.

\subsection{Decoder}

\noindent Rich ASPP semantics are bilinearly upsampled from 32$\times$32$\times$256 to 128$\times$128$\times$256 feature maps and concatenated with low-level entry flow features to resolve them in the following convolutions. This is similar to the approach used in other encoder-decoder architectures\cite{chen2017rethinking, chen2018encoder, yang2018low, badrinarayanan2017segnet, mao2016image}. There are two residual concatenations with low-level features from the entry flow; rather than one, so that semantics can be resolved from the first concatenation before being resolved into fine spatial information after the second. 

To prevent rich semantic information from being overwhelmed by low-level features, feature sizes are chosen so that, disregarding spatial dimensions, the same number of low-level and high-level features enter the decoder. The optimal ratio is unknown; however, this paper and Google's work on semantic segmentation\cite{chen2017rethinking, chen2018encoder} establish that a 1:1 ratio works for multiple domains. The resolved features are transpositionaly convoluted back to the size of the original image; rather than resolved at that scale, to introduce another bottleneck. This forces the network to work harder to develop meaningful features to resolve recovered micrographs from in the final convolutions.

\section{Training}\label{sec_training}

\begin{figure}[tbp]
\centering
\includegraphics[trim={0.3cm 0.5cm 0 0},width=0.48\textwidth]{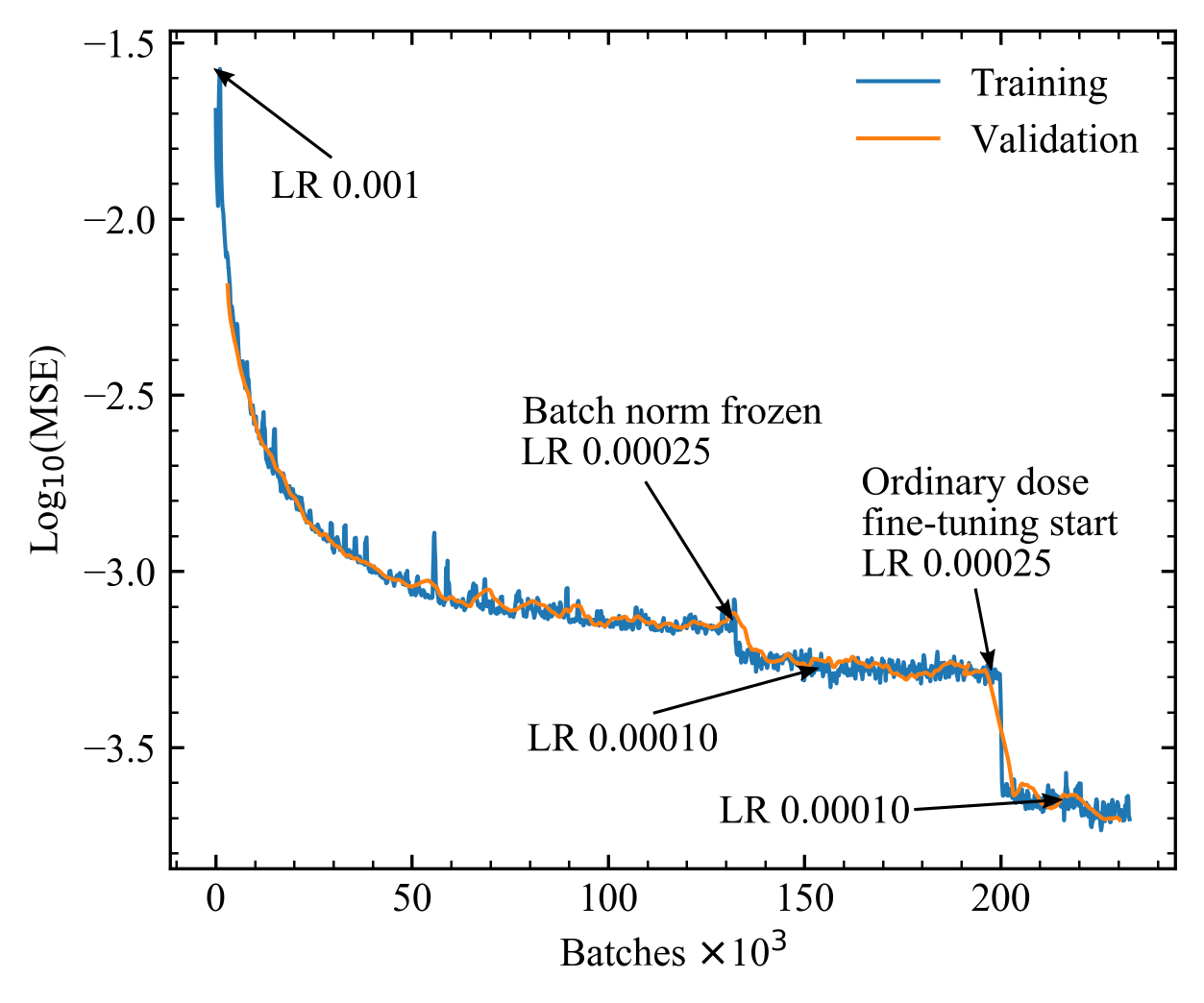}
\caption{ Mean squared error (MSE) losses of our neural network during training on low dose ($\ll 300$ counts ppx) and fine-tuning for ordinary doses (200-2500 counts ppx). Learning rates (LRs) and the freezing of batch normalization are annotated. Validation losses were calculated using 1 validation example after every 5 training batches.}
\label{learning_curves}
\end{figure}

\noindent In this section, we discuss training with the TensorFlow\cite{abadi2016tensorflow} deep learning framework. Training was performed using ADAM\cite{kingma2014adam} optimized synchronous stochastic gradient decent\cite{chen2016revisiting} with 1 replica network on each of 2 Nvidia GTX 1080 Ti GPUs. 

\subsection{Data Pipeline}\label{data_pipeline}
\noindent A new dataset of 17267 2048$\times$2048 high-quality electron micrographs saved to University of Warwick data servers over several years was collated for training. Here, high-quality refers to 2048$\times$2048 micrographs with mean counts per pixel above 2500. The dataset was collated from individual micrographs made by dozens of scientists working on hundreds of projects and therefore has a diverse constitution. It has been made publicly available as a set of TIFFs\cite{adobe1992tiff} via \url{https://github.com/Jeffrey-Ede/Electron-Micrograph-Denoiser}. 

The dataset was split into 11350 training, 2431 validation and 3486 test micrographs. For training, each micrograph was downsized by a factor of 2 using area interpolation to 1024$\times$1024. This increased mean counts per pixels above 10000, corresponding to signal-to-noise ratios above 100:1. Next, 512$\times$512 crops were taken at random positions and subject to a random combination of flips and 90$\degree$ rotations to augment the dataset by a factor of 8. Each crop was then linearly transformed to have values between 0.0 and 1.0.

To train the network for low doses, Poisson noise was applied to each crop after scaling by a number sampled from an exponential distribution with probability density function (PDF)
\begin{equation}
f\left(x,\frac{1}{\beta}\right) = \frac{1}{\beta}\exp\left(-\frac{x}{\beta}\right).
\end{equation}
We chose $\beta = 75.0$ and offset the numbers sampled by 25.0 before scaling crops with them. These numbers and distribution are arbitrary and were chosen to expose the network to a continuous range of noise levels where most are very noisy. After noise application, ground truth crops were scaled to have the same means as their noisy counterparts.

After being trained for low-dose applications, the network was fine-tuned for ordinary doses by training it on crops scaled by numbers uniformly distributed between 200 and 2500.

\subsection{Learning Policy}

\noindent In this subsection, we discuss our training hyperparameters and learning protocol for the learning curve shown in fig.~\ref{learning_curves}.

\vspace{\extraspace}
\noindent \textbf{Loss metric:} Our network was trained to minimize the Huberised\cite{huber1964robust} scaled mean squared error (MSE) between denoised and high-quality crops:

\begin{equation}
L = 
\begin{cases}
      1000\, \textit{MSE}, & 1000\, \textit{MSE} < 1.0 \\
      \left(1000\, \textit{MSE}\right)^{\frac{1}{2}}, & 1000\, \textit{MSE} \ge 1.0
\end{cases}
\end{equation}

The loss is Huberized to prevent the network from being too disturbed by batches with especially noisy crops. To surpass our low-dose performance benchmarks, our network had to achieve a MSE lower than $7.5\times10^{-4}$, as tabulated in table~\ref{results_summary}. Consequently, MSEs were scaled by 1000 to limit trainable parameter perturbations by MSEs larger than $1.0\times10^{-3}$. More subtly, this also increased our network's effective learning rate by a factor of 1000.

Our network was trained without clipping its outputs between 0.0 and 1.0. Clipping is only applied optionally during inference. Nevertheless, as clipping is desirable in most applications, all performance statistics; including losses during training, are reported for clipped outputs.

\vspace{\extraspace}
\noindent \textbf{Batch normalization}: Batch normalization layers from \cite{ioffe2015batch} were trained with a decay rate of 0.999 for 134108 batches. Batch normalization was then frozen for the rest of training. Batch normalization significantly reduced grid-like increases in error in our output images. As a result, it was not frozen until the instability introduced by varying batch normalization parameters noticeably limited convergence.

\vspace{\extraspace}
\noindent \textbf{Optimizer:} ADAM\cite{kingma2014adam} optimization was used with a stepped learning rate. For the low dose version of the network, we used a learning rate of 1.0$\times$10$^{-3}$ for 134108 batches, 2.5$\times$10$^{-4}$ for another 17713 batches and then 1.0$\times$10$^{-4}$ for 46690 batches. The network was then fine-tuned for ordinary doses using a learning rate of 2.5$\times$10$^{-4}$ for 16773 batches, then 1.0$\times$10$^{-4}$ for 17562 batches. The unusual numbers of batches are a result of learning rates being adjusted after saving variables; something that happened at wall clock times.

We found the recommended\cite{kingma2014adam,abadi2016tensorflow} ADAM decay rate for the 1st moment of the momentum, $\beta_1 = 0.9$, to be too high and chose $\beta_1 = 0.5$ instead. This lower $\beta_1$ made training more resistant to varying noise levels in batches. Our experiments with $\beta_1$ are discussed in section~\ref{additional_experiments}.

\vspace{\extraspace}
\noindent \textbf{Regularization:} L2 regularization\cite{kukavcka2017regularization} was applied by adding 5$\times$10$^{-5}$ times the quadrature sum of all trainable variables to the loss function. This prevented trainable parameters growing unbounded, decreasing their ability to learn in proportion\cite{salimans2016weight}. Importantly, this ensures that our network will continue to learn effectively if it is fine-tuned or given additional training. We did not perform an extensive search for our regularization rate and think that 5$\times$10$^{-5}$ may be too high.

\vspace{\extraspace}
\noindent \textbf{Activation:} All neurons are ReLU6\cite{krizhevsky2010convolutional} activated. Other activations are discussed in section~\ref{additional_experiments}.

\vspace{\extraspace}
\noindent \textbf{Initialization:} All weights were Xavier\cite{glorot2010understanding} initialized. Biases were zero initialized.

\section{Performance}\label{sec_performance}

\noindent In this section, our network's MSE and structural similarity index (SSIM) performance is benchmarked other methods' and its mean error per pixel is mapped. We also present example applications of our network to noisy electron microscopy and scanning transmission electron microscopy (STEM) images.

\subsection{Benchmarking}

\begin{figure*}[tbp]
\centering
\includegraphics[width=0.96\textwidth]{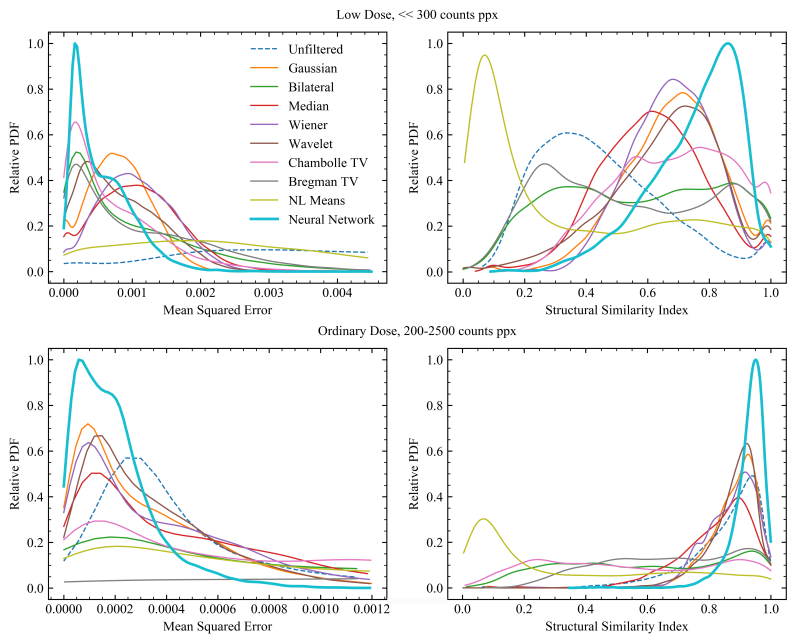}
\caption{ Gaussian kernel density estimated mean squared error (MSE) and structural similarity index (SSIM) probability density functions (PDFs) for denoising methods applied to 20000 instances of Poisson noise. To ease comparison, the highest values in each MSE and SSIM PDF set has been scaled to 1.0. Only the starts of MSE PDFs are shown.}
\label{kdes}
\end{figure*}

\noindent To benchmark our network's performance, we applied it and 9 popular denoising methods to 20000 instances of Poisson noise applied to 512$\times$512 crops from test set micrographs using the method in section~\ref{data_pipeline}. This benchmarking is done for the low-dose version of our network and the version fine-tuned for ordinary doses. Implementation details for each denoising methods follow.

\begin{table*}[]
\begin{tabular*}{\textwidth}{@{\extracolsep{\fill}}l|cccc|cccc}
\hline
                            & \multicolumn{4}{c|}{Low Dose, $\ll$ 300 counts ppx}              & \multicolumn{4}{c}{Ordinary Dose, 200-2500 counts ppx}                        \\
\multicolumn{1}{c|}{}       & \multicolumn{2}{c}{MSE $\left(\times10^{-3}\right)$} & \multicolumn{2}{c|}{SSIM} & \multicolumn{2}{c}{MSE $\left(\times10^{-3}\right)$} & \multicolumn{2}{c}{SSIM} \\ \hline
Method & Mean         & \multicolumn{1}{c}{Std Dev}      & Mean       & Std Dev      & Mean       & \multicolumn{1}{c}{Std Dev}       & Mean      & Std Dev      \\ \hline
Unfiltered                  & 4.357        & 2.558                             & 0.454      & 0.208        &      0.508      &                  0.682                  &      0.850     &     0.123         \\
Gaussian                    & 0.816        & 0.452                             & 0.685      & 0.159        &       0.344     &                   0.334                 &      0.878     &       0.087       \\
Bilateral                   & 1.025        & 1.152                             & 0.574      & 0.261        &       1.243     &                  1.392                  &      0.600     &       0.271       \\
Median                      & 1.083        & 0.618                             & 0.618      & 0.171        &      0.507      &                    0.512                &      0.821     &        0.126      \\
Wiener          & 1.068                 & 0.546                   & 0.681      & \textbf{0.137}         &     0.402       &                 0.389                   &     0.870       &       0.085       \\
Wavelet                     & 0.832        & 0.58                              & 0.657      & 0.186        &      0.357      &                  0.312                  &      0.875     &       0.085       \\
Chambolle TV                & 0.746        & 0.725                             & 0.686      & 0.192        &      0.901      &                0.909                    &      0.674     &       0.217       \\
Bregman TV      & 1.109                 & 1.031                   & 0.544      & 0.268          &      4.074      &                   3.025                 &      0.348     &      0.312        \\
NL means        & 2.924                 & 2.338                   & 0.357      & 0.315         &       1.403     &                1.266                    &     0.545      &     0.281           \\ \hline
Neural network  & \textbf{0.562}                   & \textbf{0.449}                     & \textbf{0.752}        & 0.147          &    \textbf{0.201}        &            \textbf{0.169}                        &     \textbf{0.926}      &        \textbf{0.057}      \\ \hline
\end{tabular*}
\caption{Means and standard deviations of mean squared error (MSE) and structural similarity index (SSIM) test set performances of denoising methods for 20000 instances of Poisson noise. Noise removal methods were implemented with default parameters, as described in the main text. TV - total variation. NL - non-local.}
\label{results_summary}
\end{table*}

\vspace{\extraspace}
\noindent \textbf{Unfiltered:} For reference, MSE and SSIM statistics were collected for noisy crops without any denoising method being applied to recover them.

\vspace{\extraspace}
\noindent \textbf{Gaussian filter:} OpenCV\cite{opencv_library} default implementation of Gaussian blurring for a 3$\times$3 kernel.

\vspace{\extraspace}
\noindent \textbf{Bilateral filter:} OpenCV\cite{opencv_library} implemented bilateral filtering\cite{tomasi1998bilateral}. We used a default 9 pixel neighborhoods with radiometric and spatial scales of 75. These scales are a compromise between small scales; less than 10, that have little effect and large scales; more than 150, that cartoonize images.

\vspace{\extraspace}
\noindent \textbf{Median filter:} OpenCV\cite{opencv_library} implemented 3$\times$3 median filter.

\vspace{\extraspace}
\noindent \textbf{Wiener:} Scipy\cite{scipy} implementation of Wiener filtering.

\vspace{\extraspace}
\noindent \textbf{Wavelet denoising:} Scikit-image\cite{van2014scikit} implementation of BayesShrink adaptive wavelet soft-thresholding\cite{chang2000adaptive}. The noise standard deviation used to compute wavelet detail coefficient thresholds was estimated from the data using the method in \cite{donoho1994ideal}.

\vspace{\extraspace}
\noindent \textbf{Chambolle total variation denoising:} Scikit-image\cite{van2014scikit} implementation of Chambolle's iterative total-variation denoising algorithm\cite{chambolle2004algorithm}. We used the scikit-image default denoising weight of 0.1 and ran the algorithm until either the fractional change in the algorithm's cost function was less than $2.0\times 10^{-4}$ or it reached 200 iterations.

\vspace{\extraspace}
\noindent \textbf{Bregman total variation denoising:} Scikit-image\cite{van2014scikit} implementation of split Bregman anisotropic total variation denoising\cite{goldstein2009split, getreuer2012rudin}. We used a denoising weight of 0.1 and ran the algorithm until either the fractional change in the algorithm's cost function was less than $2.0\times 10^{-4}$ or it reached 200 iterations

\vspace{\extraspace}
\noindent \textbf{Non-local means}: Scikit-image\cite{van2014scikit} implementation of texture-preserving non-local means denoising\cite{buades2005jm}.

MSE and SSIM performances were divided into 200 equispaced bins in [0.0, 1.2] $\times$ $10^{-3}$ and [0.0, 1.0], respectively, for both low and ordinary doses. Performance PDFs shown in fig.~\ref{kdes} are Gaussian kernel density estimated\cite{turlach1993bandwidth, bashtannyk2001bandwidth} (KDE) from these. KDE bandwidths were found using Scott's Rule\cite{scott2015multivariate}. To ease comparison, PDFs for each set of performance statistics are scaled so the highest value in each set is 1.0. The scale factors are 6.07$\times$10$^{-4}$ for low dose MSE, 2.99$\times$10$^{-1}$ for low dose SSIM, 3.03$\times$10$^{-4}$ for ordinary dose MSE and 9.17$\times$10$^{-2}$ for ordinary dose SSIM.

Performance statistics are also summarized in table~\ref{results_summary}. This shows that our network outperforms the mean MSE and SSIM performance of every other method for both low and ordinary doses. Its MSE variance is also lower for both low and ordinary doses.

\begin{figure*}[tbp]
\centering
\includegraphics[width=0.95\textwidth,resolution=600]{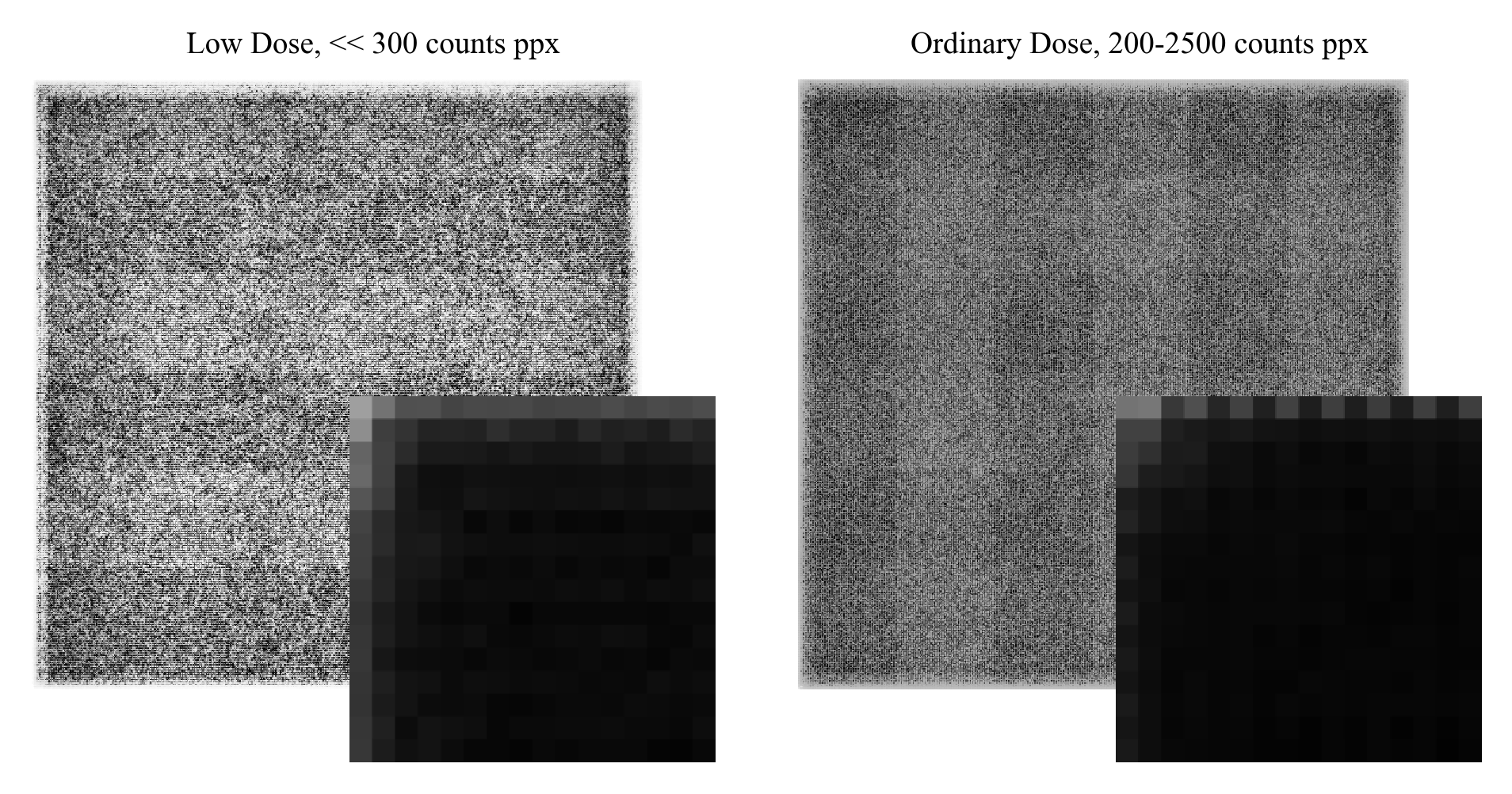}
\caption{ Mean absolute errors of our low and ordinary dose networks' 512$\times$512 outputs for 20000 instances of Poisson noise. Contrast limited adaptive histogram equalization\cite{zuiderveld1994contrast} has been used to massively increase contrast, revealing grid-like error variation. Subplots show the top-left 16$\times$16 pixels' mean absolute errors unadjusted. Variations are small and errors are close to the minimum everywhere, except at the edges where they are higher. Low dose errors are in [0.0169, 0.0320]; ordinary dose errors are in [0.0098, 0.0272]. }
\label{abs_err}
\end{figure*}

\subsection{Network Error}\label{Characterization}

\noindent Mean absolute errors for each pixel of our network's output are shown for low and ordinary doses in fig.~\ref{abs_err}. The errors are uniform almost everywhere, except at the edges where they are higher. For low and ordinary doses, the mean absolute errors per pixel are 0.0177 and 0.0102, respectively.

Small, grid-like variations in absolute error are revealed by contrasted limited adaptive histogram equalization\cite{zuiderveld1994contrast} in fig.~\ref{abs_err}. These variations are common in deep learning and are often associated with transpositional convolutions. Consequently, some authors\cite{odena2016deconvolution} have recommended their replacement with bilinear upsampling then convolution. We tried this; however, it only made the errors less grid-like. 

Instead, we found batch normalization to be the best way to reduce structured error variation. This is demonstrated by errors being more grid-like for the ordinary dose version of our network, which was trained for longer after batch normalization was frozen. Consequently, batch normalization was not frozen until the instability introduced by its trainable variables noticeably limited convergence. 

\subsection{Example Usage}

\begin{figure*}[tbp]
\centering
\includegraphics[width=0.9\textwidth,resolution=600]{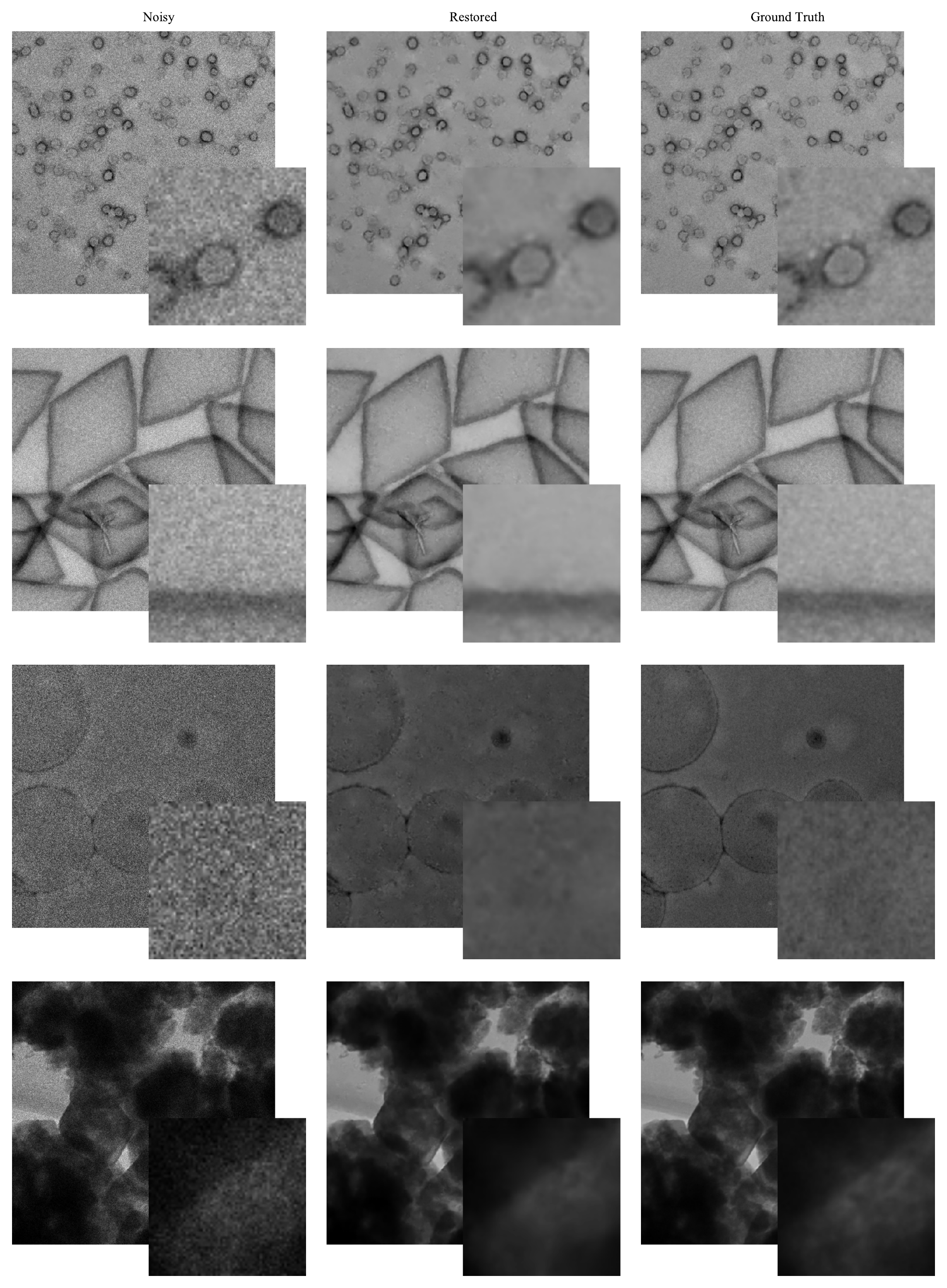}
\caption{ Example applications of the noise-removal network to instances of Poisson noise applied to 512$\times$512 crops from high-quality micrographs. Enlarged 64$\times$64 regions from the top left of each crop are shown to ease comparison. }
\label{TEM_example1}
\end{figure*}

\begin{figure*}[tbp]
\centering
\includegraphics[width=0.9\textwidth,resolution=600]{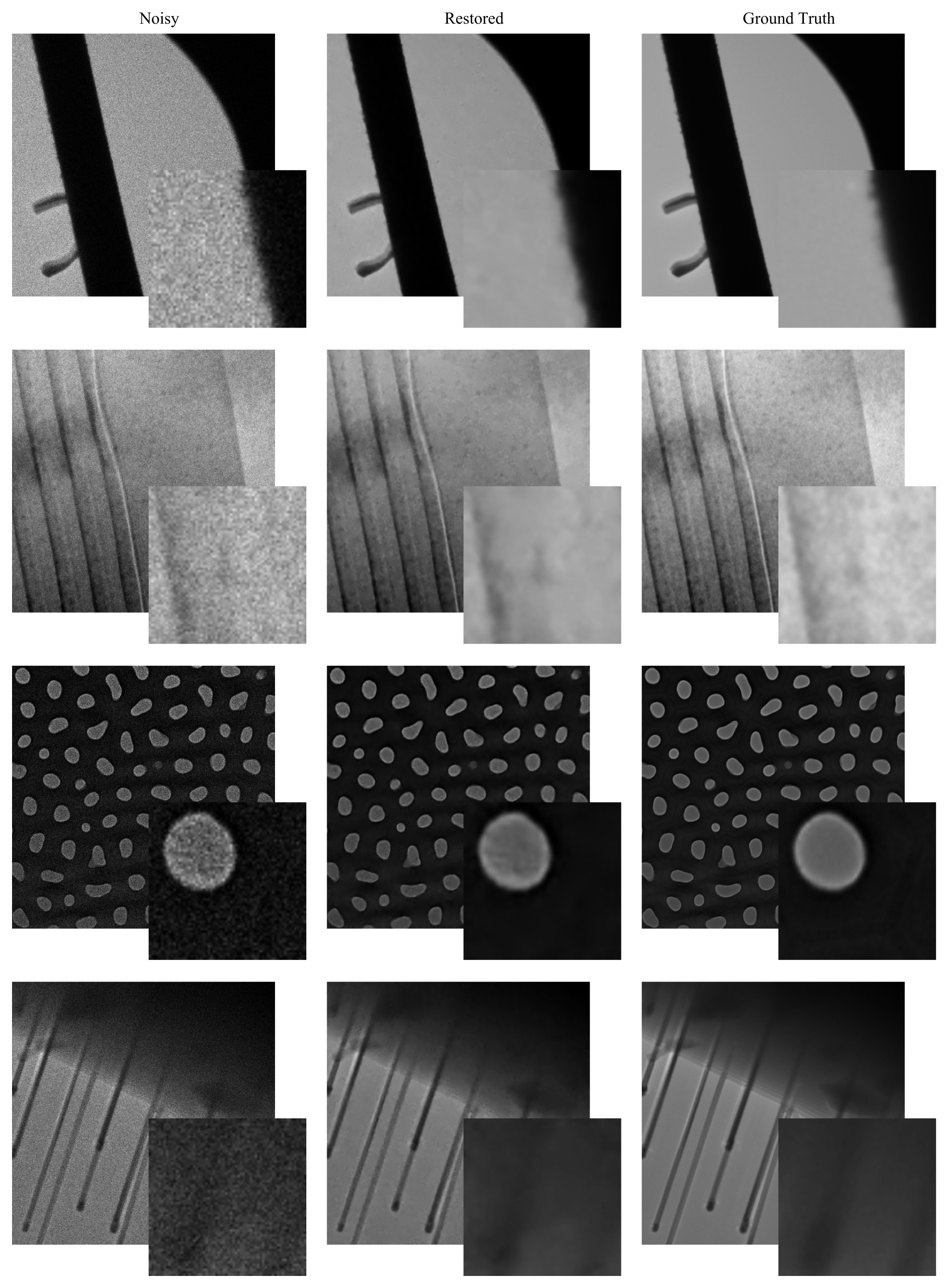}
\caption{ More example applications of the noise-removal network to instances of Poisson noise applied to 512$\times$512 crops from high-quality micrographs. Enlarged 64$\times$64 regions from the top left of each crop are shown to ease comparison. }
\label{TEM_example2}
\end{figure*}

\begin{figure*}[tbp]
\centering
\includegraphics[width=0.95\textwidth,resolution=600]{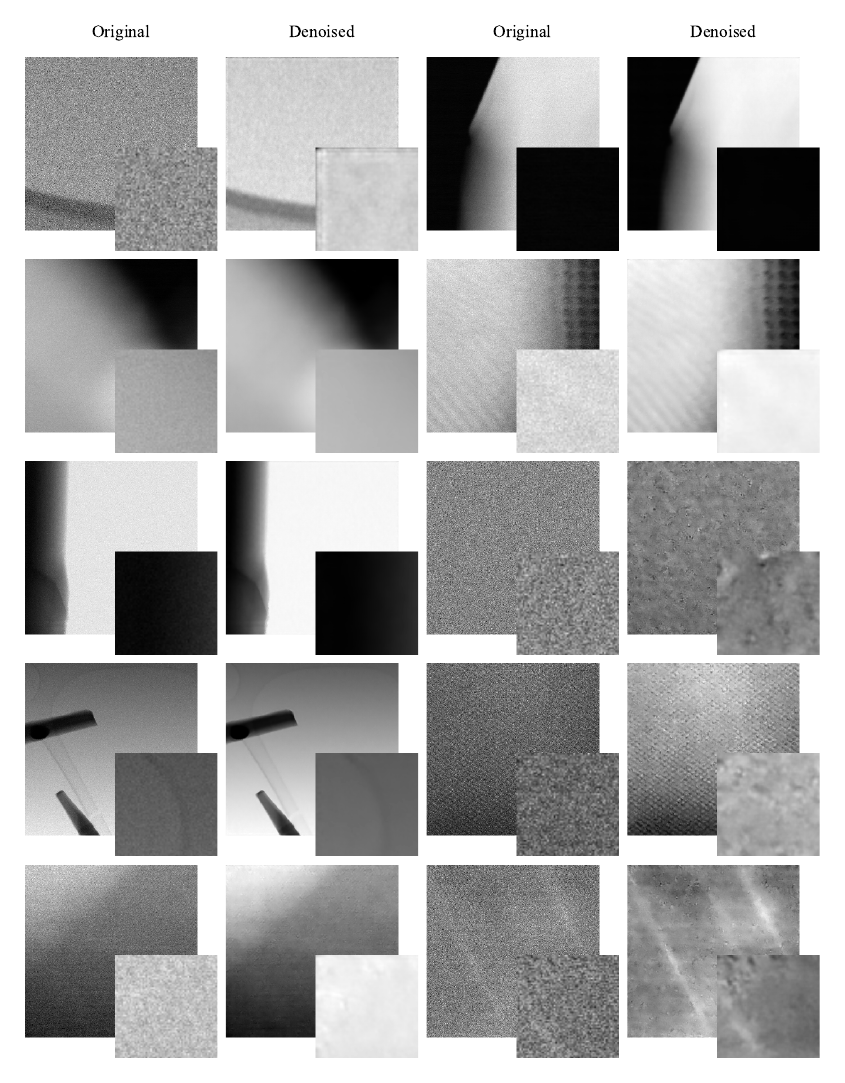}
\caption{ Example applications of our neural network to 512$\times$512 crops from scanning transmission electron micrographs. Enlarged 64$\times$64 regions from the top left of each crop are shown to ease comparison. Our network has not been trained for this domain. }
\label{STEM_example1}
\end{figure*}

\noindent We provide several example applications of our low-dose network to 512$\times$512 crops from high-quality electron micrographs that we applied noise to in fig~\ref{TEM_example1} and fig.~\ref{TEM_example2}. Several example applications of our low-dose network to 512$\times$512 crops from scanning transmission electron micrography (STEM) images are also presented in fig.~\ref{STEM_example1}. Our network has not been trained for STEM so this demonstrates its ability to generalize to other domains. 

Our neural network is designed to be simple and easy to use. An example of our network being loaded once and used for inference multiple times in python is

\begin{lstlisting}[language=python]
>>> from denoiser import Denoiser
>>> noise_remover = Denoiser()
>>> restored_img1 = noise_remover.denoise(img1)
>>> restored_img2 = noise_remover.denoise(img2)
\end{lstlisting}

Under the hood, our program divides images larger than 512$\times$512 into slightly overlapping 512$\times$512 crops that can be processed by our network. This gives higher accuracy than using non-overlapping crops as our network has much higher errors for the couple of pixels near the edges of its outputs. Reflection padding is applied to images before cropping to reduce errors at their edges after restoration. Users can customize the amount overlap, padding and many other options or use default values we have chosen.

We speed-tested our network by applying it to 20000 512$\times$512 images with 1 external GTX 1080 Ti GPU and 1 thread of a i7-6700 processor. It has a mean batch size 1 (worst case) inference time of 77.0 ms. It also takes a few seconds to load before it is ready for repeated use.




\section{Additional Experiments}\label{additional_experiments}
\noindent As part of development, we experimented with multiple architectures and their learning policies. Initially, we experimented with shallower architectures similar to \cite{yang2018low, mao2016image} and \cite{zhang2017beyond}; however, these struggled to meet Chambolle's low-dose benchmark in table~\ref{results_summary}. Consequently, we switched to the deeper Xception-based architecture presented in this paper. In this section we present some of the experiments we performed to fine-tune it.

\begin{figure}[tbp]
\centering
\includegraphics[trim={0.3cm 0.5cm 0 0},width=0.48\textwidth]{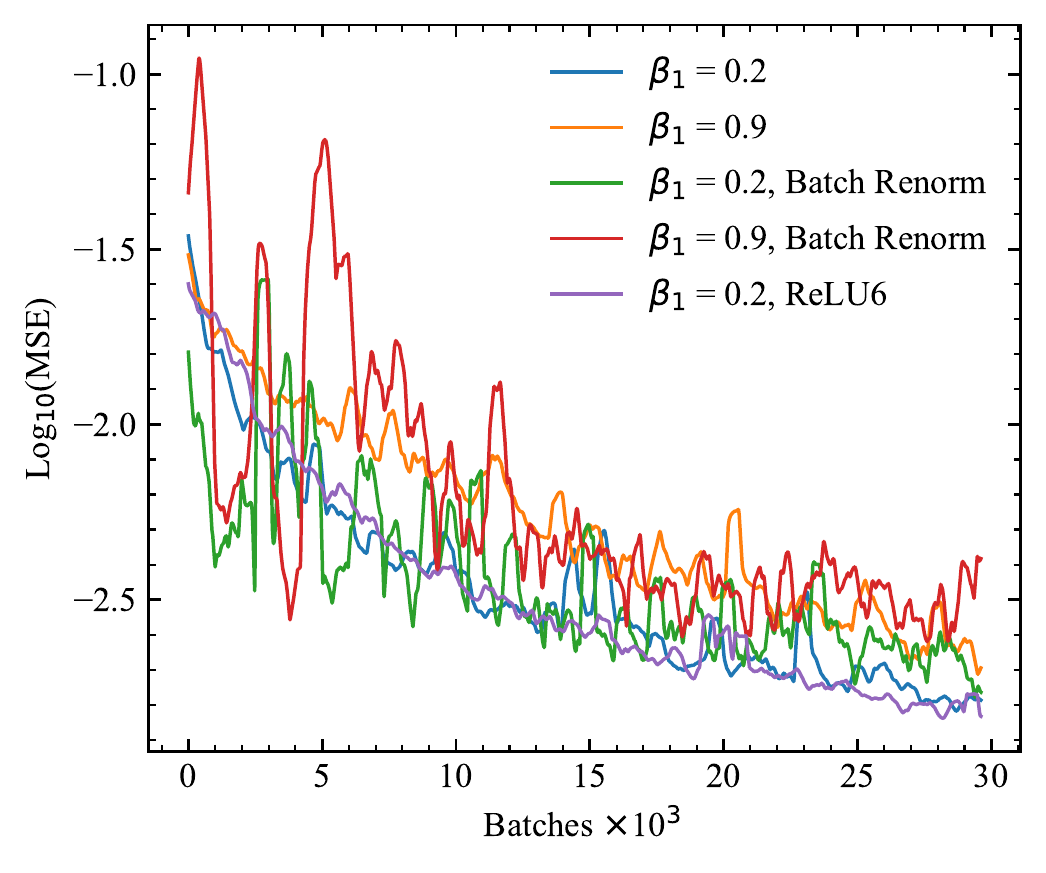}
\caption{Batch size 2 mean squared error (MSE) learning curves for various training hyperparameters and activation functions. Training is most stable for ADAM\cite{kingma2014adam} $\beta_1 = 0.2$ without batch renormalization\cite{ioffe2017batch} and with ReLU6\cite{krizhevsky2010convolutional}; rather than ReLU\cite{nair2010rectified}, activation.}
\label{initial_learning_curves}
\end{figure}

\begin{figure}[tbp]
\centering
\includegraphics[trim={0.3cm 0.5cm 0 0},width=0.48\textwidth]{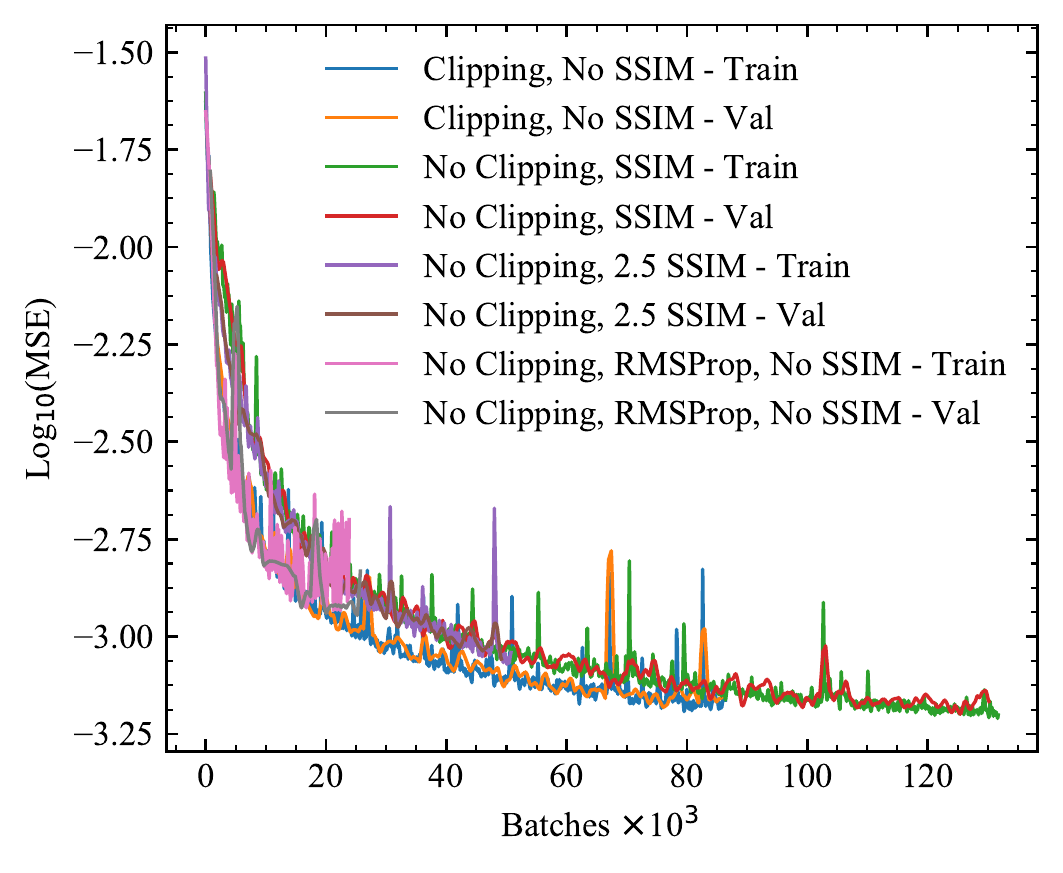}
\caption{Batch size 10 mean squared error (MSE) learning curves for training with and without clipping,  extra structural similarity index\cite{wang2004image} (SSIM) based losses and RMSProp\cite{hinton2012neural} rather than ADAM \cite{kingma2014adam} optimization.}
\label{learning_curves_10}
\end{figure}

\subsection{Batch Size 2}

\noindent Initial experiments with a batch size of 2 are summarized in fig.~\ref{initial_learning_curves}. This includes experiments with different decay rates for the 1st moment of the momentum, $\beta_1$, used by the ADAM solver. Our results show that training converges faster and more stably for $\beta_1=0.2$ than the recommended\cite{kingma2014adam, abadi2016tensorflow} $\beta_1=0.9$. $\beta_1=0.9$ being too high seems to be a theme in high resolution, low batch size applications e.g. \cite{wang2017high}.

Our batch size 2 experiments also revealed that training is faster and more stable with ReLU6\cite{krizhevsky2010convolutional} than with ReLU\cite{nair2010rectified} activation. We think this is because ReLU6 is especially effective in the decoder where it does not allow sparse activations to grow unbounded. Only ReLU-based activation functions were considered as the non-saturation of their gradients accelerates the convergence of stochastic gradient descent\cite{krizhevsky2012imagenet}.

Batch renormalization\cite{ioffe2017batch} was trailed as our limited hardware made large batch sizes prohibitive. We also wanted to test how well it would work with batch size 2 as we were unable to find quantitative information for this batch size. It is ineffective: Fig.~\ref{initial_learning_curves} shows that the extra trainable variables it introduces destabilize training and reduce convergence. This is not surprising as it is not designed for batches this small and can take time to become effective. Additionally, instability introduced by applying different amounts noise to training examples may be adding to the inherent batch renormalization instability, making it much worse than it might be for other optimizations.

\subsection{Batch Size 10}

\noindent We tried adding multiples of the distance from the maximum SSIM, $1.0-\textit{SSIM}$, to the training loss to optimize our network for an additional metric. As shown by fig.~\ref{learning_curves_10}, adding different multiples does not have a significant effect on the rate of MSE convergence. Regardless, we decided against adding this loss as SSIMs are weighted to measure a human perception of quality and we did not want to introduce this slight bias. Nevertheless, we have shown that additionally optimizing the SSIM does not significantly affect training.

Outputs were clipped between 0.0 and 1.0 before calculating training losses in all of our batch size 2 and the first of our batch size 10 experiments shown in fig.~\ref{learning_curves_10}. Removing this clipping significantly decreases the rate of convergence. Nonetheless, we decided to remove clipping as the network was especially prone to producing artifacts close to 0.0 and 1.0. This ensured that these artifacts were not present in the fully trained network's outputs.

Following the success of lower $\beta_1$ with the ADAM optimizer, we tried other momentum-based optimizers. The start of a learning curve for simple RMSProp\cite{hinton2012neural} is shown in fig.~\ref{learning_curves_10}. It shows that training is significantly less stable and has a lower rate of convergence. We also experimented with Nesterov-accelerated gradient descent\cite{sutskever2013importance,nesterov1983method} with similar results.

\section{Summary}
\begin{itemize}
\item We have developed a deep neural network for electron micrograph denoising and shown that it outperforms existing methods for low and ordinary electron doses. 
\item Fully trained versions of our network have been made available for low and ordinary doses with example usage: \url{https://github.com/Jeffrey-Ede/Electron-Micrograph-Denoiser}
\item A new dataset of 17267 2048$\times$2048 high-quality micrographs collected to train our network has been made publicly available.
\item Example applications of our network to noisy electron micrographs are presented. We also present example application to STEM images to show that our network can generalize to other domains.
\item Our network architecture, training hyperparameters and learning protocols are detailed. In addition, details of several of our initial architecture and learning policy experiments are presented.
\end{itemize}

\bibliographystyle{ieeetr}
\bibliography{bibliography}
\end{document}